\documentclass[letterpaper, 10 pt, conference]{ieeeconf} 
\IEEEoverridecommandlockouts           
\usepackage{cite}
\usepackage{amsmath,amssymb,amsfonts}
\usepackage{algorithmic}
\usepackage{graphicx}
\usepackage{textcomp}

\usepackage{multirow}
\usepackage{tabularx} 
\usepackage{caption}

\usepackage{xcolor}

\title{\LARGE \bf
CFDNet: A Generalizable Foggy Stereo Matching Network with Contrastive Feature Distillation
}

\author{Zihua Liu, Yizhou Li and Masatoshi Okutomi.
\thanks{All authors are with the Department of Systems and Control Engineering, School of Engineering,
Tokyo Institute of Technology. Tokyo, Japan.
{\tt\small \{zliu,yli,mxo\}@ok.sc.e.titech.ac.jp}}}
\begin{document}
\maketitle
\thispagestyle{empty}
\pagestyle{empty}

\begin{abstract}
Stereo matching under foggy scenes remains a challenging task since the scattering effect degrades the visibility and results in less distinctive features for dense correspondence matching. While some previous learning-based methods integrated a physical scattering function for simultaneous stereo-matching and dehazing, simply removing fog might not aid depth estimation because the fog itself can provide crucial depth cues. In this work, we introduce a framework based on contrastive feature distillation~(CFD). This strategy combines feature distillation from merged clean-fog features with contrastive learning, ensuring balanced dependence on fog depth hints and clean matching features. This framework helps to enhance model generalization across both clean and foggy environments. Comprehensive experiments on synthetic and real-world datasets affirm the superior strength and adaptability of our method.

\end{abstract}

\section{Introduction}

Stereo matching is a crucial task, where depth is inferred using the disparity between left and right images. Given the affordability of stereo cameras, this technique is popular in fields like Augmented Reality (AR) and autonomous driving. However, in practical situations, especially in the context of autonomous driving, adverse weather conditions such as fog or haze are common. These conditions significantly reduce image clarity, posing challenges for stereo matching. Notably, learning-based stereo-matching networks, typically trained on clean images, exhibit apparent performance degradation in foggy conditions, as illustrated in Fig. \ref{fig:FogDegragation}.

Prior researches~\cite{BidNet,HazeStereoRefinement,CAT} employ networks to mitigate fog's effects on matching by reconstructing the clean image using the atmospheric scattering model~\cite{VisionInBadWeather,SingleImageVisibility,FogVoluming}. These networks simultaneously predict two fog-related hyperparameters: the transmission parameter \textit{T} and the airlight $L_{\infty}$, while estimating disparity~\cite{FoggyStereo,song2020simultaneous}.
Estimating scattering hyperparameters is challenging with the exponential relationship between transmission and depth. Small errors in transmission estimation can result in significant disparity errors, making training unstable and less generalizable in real-world scenarios.

In contrast to physical prior-based stereo-matching methods for foggy conditions that combine image dehazing, this paper emphasizes enhancing matching feature representations from both clean and foggy images to boost stereo matching across varying conditions. We propose a contrastive feature distillation~(CFD) strategy and a network named \textit{CFDNet}. Surprisingly, we discovered that fog does not invariably degrade stereo matching. The fog's thickness can offer valuable depth cues, where we demonstrate its effectiveness in enhancing matching features significantly with our contrastive feature distillation. Our contributions are summarized as follows:
\begin{figure}[!t]
    \centering
\includegraphics[width=1.0\linewidth]{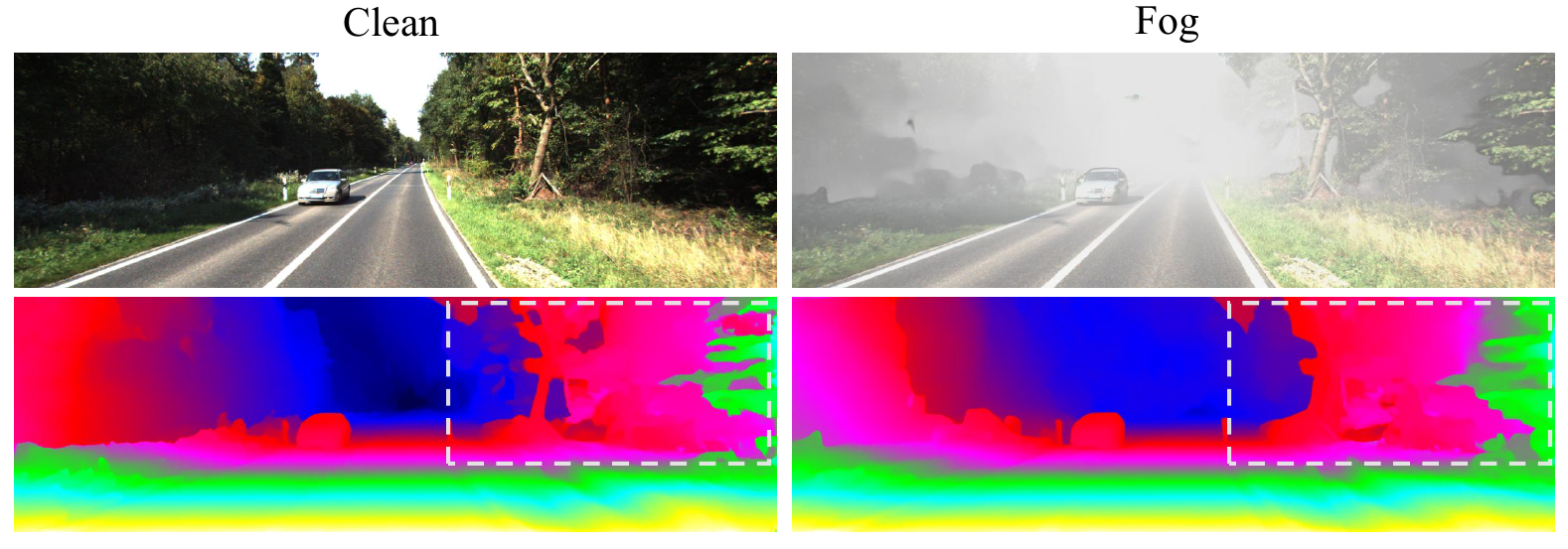}
    \caption{Clean and foggy left images from stereo image pairs (top), and estimated disparity maps from each pair with state-of-the-art RaftStereo\cite{RaftStereo} (bottom).} 
    \label{fig:FogDegragation}
    \vspace{-5mm}
\end{figure}
\begin{itemize}
    \item Contrastive Feature Distillation: We incorporate contrastive learning into feature distillation, ensuring a balanced distilled feature representation between clean and foggy domains. Based on this, we introduce a teacher-student model that optimizes performance across both domains by harmonizing reliance on fog depth cues with clean matching features.
    \item Attentive Feature Converter: We propose an attentive feature converter with stacked pixel and channel-wise attention. It enables fine-grained feature fusion for the \textit{Teacher Model} and feature adaptation for \textit{Student Model} with proposed contrastive feature distillation.
    \item We conducted comprehensive experiments in both synthetic and real-world settings, utilizing datasets including SceneFlow~\cite{dispnetc}, KITTI~\cite{KITTIDatset}, and PixelAccurateDepeth~\cite{PixelAccurateDepth}. Proposed \textit{CFDNet} surpasses leading methods in both foggy and clean environments, underlining its strong efficacy and applicability.
\end{itemize}

\section{Related Work}
\subsection{Stereo Matching}
Stereo Matching has been an active field and well-studied for decades. Traditional approaches\cite{AdaptiveWindow,SGM,PatchMatchStereo,GraphCutStereo,StereoBriefPropagation} employ handcrafted patterns for local correspondence matching and exploit spatial context for global optimization. Recently, many learning-based methods\cite{dispnetc,psmnet, RaftStereo,ganet,gwcnet, IGEVStereo} use deep networks to learn discriminative matching features and rely on a cost volume architecture with 2D or 3D convolution aggregation for effective stereo matching. Although these methods have made great progress, since they are often trained on clean images only, they exhibit severe performance degradation when encountering foggy images, posing challenges for real-world applications like autonomous driving.

\subsection{Stereo Matching in Foggy Conditions.}
Enhancing the adaptability of stereo-matching algorithms to various scenarios, especially foggy conditions, has drawn significant research interest.
\cite{gated2depth} employs a CMOS gated camera alongside a standard stereo camera, aiding depth estimation under severe weather.
CATS\cite{CAT} provides a color and thermal stereo benchmark by applying thermal-thermal matching to improve the robustness of depth estimation in foggy scenes.
Caraffa et al.\cite{caraffa2012stereo} present an MRF model designed for simultaneous stereo matching and dehazing by iterative optimization using $\alpha$-expansion.
Song et al.\cite{song2018deep} pioneer a joint-learning framework, combining stereo matching and dehazing using a dual-branch structure with feature fusion. A subsequent version\cite{song2020simultaneous} adopts feature attention for a more adaptable feature integration.
Shahriar et al.\cite{negahdaripour2014improved} present an integrated solution for 3-D structure that incorporates the visual cues from both disparity and scattering.
FoggyStereo\cite{FoggyStereo} delves into depth cues that are hidden in the foggy images, formulating a foggy volume using estimated scattering coefficients to facilitate disparity estimation.
While these strategies advance performance under foggy conditions, many necessitate the estimation of scattering coefficients, which can degrade network robustness by introducing uncertainties.

\subsection{Feature Distillation and Contrastive Learning.}
Feature distillation\cite{hinton2015distilling} commonly serves to transfer knowledge from a teacher model to a student model by mimicking the soft output of the teacher. In this context, Hong et al.\cite{DistImDehaze} introduced a novel method for heterogeneous task imitation. They utilized features derived from clean images as a teaching mechanism, guiding the extraction of features from their hazy counterparts for image dehazing, achieving remarkable results. Moreover, to bridge the domain gap between fog and clean image realms, FIFO\cite{lee2022fifo} implemented a contrastive learning approach. This method compels the network to discern fog-invariant features under the constraints of a contrastive training scheme, thereby exhibiting commendable generalization in segmentation tasks. However, despite the widespread application of both contrastive and distillation learning in representation learning, their potential contributions to stereo matching, particularly under challenging foggy conditions, remain underexplored.

\section{Proposed methods}
In this section, we will provide a thorough introduction to our proposed Contrastive Feature Distillation Network~(\textit{CFDNet}): the overall architecture of our model is demonstrated in Subsection \ref{network_achitecture_chap} following the instruction of the attentive feature converter in Subsection \ref{feature_converter_chap}. Details about the \textit{Teacher Model} and the \textit{Student Model} can be found in Subsection \ref{techer_model_chap} and Subsection \ref{student_model_chap}.
\begin{figure*}[!th]
    \centering
\includegraphics[width=1.0\linewidth]{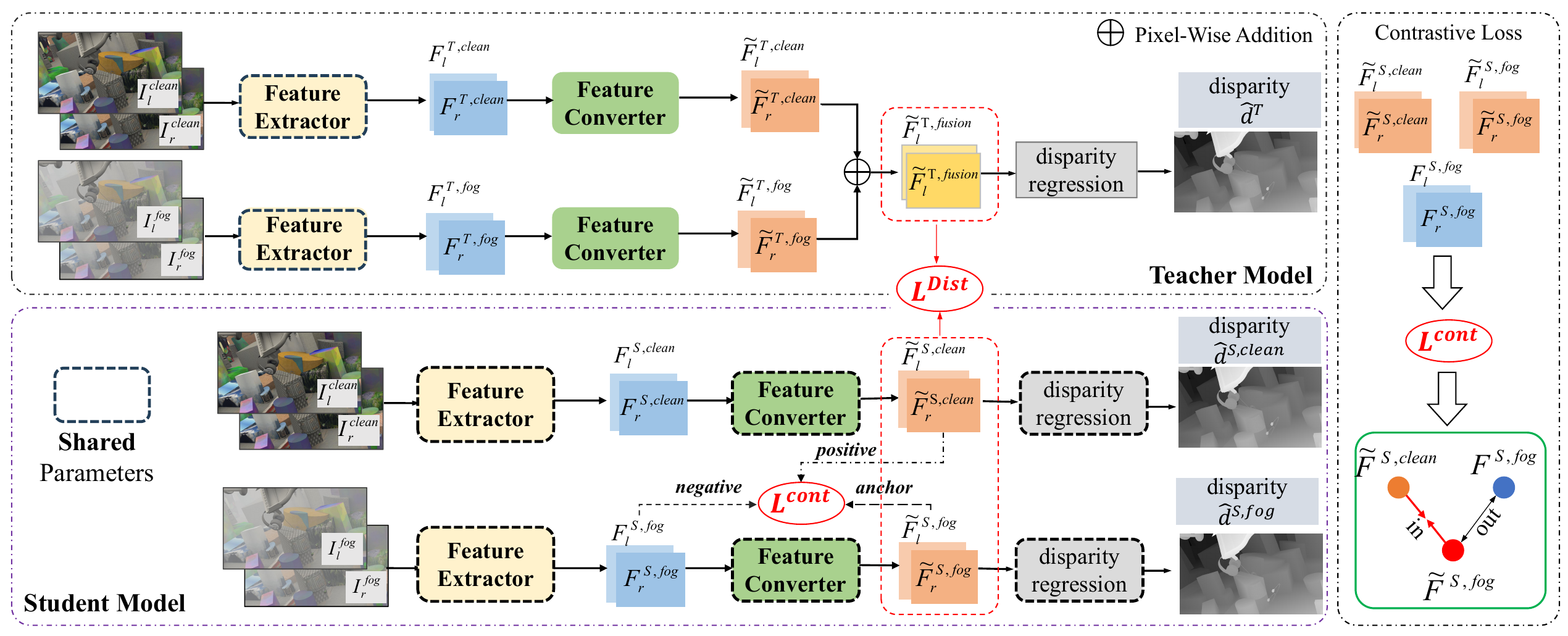}
    \caption{Overall architecture of our proposed \textit{CFDNet} with contrastive feature distillation.}
\label{fig:MainArchitecture}
\end{figure*}
\begin{figure}
    \centering
\includegraphics[width=1.0\linewidth]{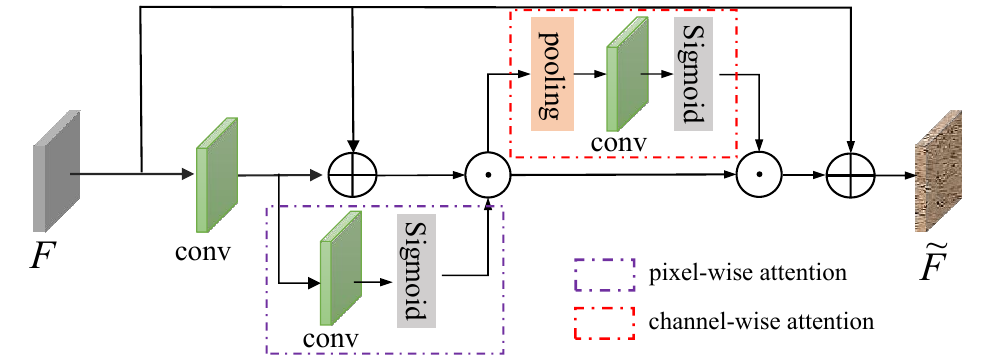}
    \caption{Attentive Feature Converter. A feature aggregation module with pixel-wise and channel-wise attention layers.}
\label{fig:feature_converter}
    \vspace{-4mm}
\end{figure}
\subsection{Network Architecture Overview}\label{network_achitecture_chap}
The overview architecture of our \textit{CFDNet} with contrastive feature distillation strategy is shown in Fig.~\ref{fig:MainArchitecture}. The whole network consists of a \textit{Teacher Model} and a \textit{Student Model}. Specifically, the \textit{Teacher Model} takes pairs of clean and foggy images as input, merging features adapted by an attentive feature converter. The \textit{Student Model}, having a similar structure, can accept single domain inputs. It benefits from the guidance of the adaptively fused features by the \textit{Teacher Model}. For robust performance in both hazy and clean domains for the \textit{Student Model}, we introduce a contrastive learning strategy that balances reliance on fog depth hints and matching features, preventing over-dependence on either one. When inference on either clean or foggy images, we utilize our pre-trained \textit{Student Model} for estimating the disparities for real-world applications.
\subsection{Attentive Feature Converter}\label{feature_converter_chap}
Motivated by recent advancements in image dehazing \cite{DCP,FFANet,DEA-Net}, discrepancies between features extracted from foggy and clean images often manifest in both spatial and channel dimensions.  Fig.~\ref{fig:feature_converter} illustrates our attentive feature converter architecture. Similar to the FFA module proposed in \cite{FFANet}, our proposed feature converter takes the initial feature 
$F$ from the feature extractor as input. It employs consecutive pixel-wise attention (PA) and channel-wise attention (CA) modules to synergistically modify features in spatial and channel dimensions. This process can be formulated as:
\begin{gather}
    \widetilde{F} = F + (F + Conv(F)) \otimes PA \otimes CA,
\end{gather}
where the $\widetilde{F}\in\mathbb{R}^{H\times W \times C}$ signifies the converted feature, while $PA$ and $CA$ represent pixel-wise and channel-wise attention map, respectively. We utilize a $3 \times 3$ convolution layer with sigmoid activation to generate a spatial attention map $PA$ of dimensions $\mathbb{R}^{H\times W \times 1}$. Similar operations are conducted for the $CA\in\mathbb{R}^{1\times 1 \times C}$, except involving a global average pooling:
\begin{gather}
PA = \sigma(Conv(F')  \\
CA = \sigma(Conv(avgpool(F')) 
\end{gather}
where the $F'$ represents the input feature and $\sigma$ stands for sigmoid function. 

As depicted in Fig.~\ref{fig:MainArchitecture}, the attentive feature converter assumes distinct roles within the Teacher and \textit{Student Model}s. In the \textit{Teacher Model}, the attentive feature converter facilitates the adaptively fused of both fog and clean features in spatial and channel domains, thereby enhancing feature representation learning. In the \textit{Student Model}, the attentive feature converter mainly focuses on alleviating the feature gap between the fog feature $F^{S,fog}$ and clean feature $F^{S,clean}$ with our contrastive feature distillation.
\subsection{Teacher~:~Feature Fusion Stereo Matching Network}\label{techer_model_chap}
A simplified RaftStereo\cite{RaftStereo} is used for disparity regression in our \textit{Teacher Model}. We first extract features $F^{T,clean}$ and $F^{T,fog}$ from paired clean and foggy stereo images by a weight-sharing feature extractor, as depicted in Fig.~\ref{fig:MainArchitecture}. Notably, though fog affects clarity, it offers depth cues beneficial for disparity regression from a mono-depth aspect. This is rooted in the inverse relationship between fog density and scene depth. Hence, in challenging situations like occlusions, foggy images can provide crucial guidance absent in clean images, while clean images provide vital structural information for matching. As both features have benefits and are complementary to each other, we adaptively fuse them using our attentive feature converter, as depicted in Fig.~\ref{fig:MainArchitecture}. This fusion is represented as:
\begin{gather}
\widetilde{F}^{T,fusion} = FC(\widetilde{F}^{T,clean}) + FC(\widetilde{F}^{T,fog}),
\end{gather}
where $\widetilde{F}^{T,fusion}$ signifies the adaptively fused feature, and $FC$ denotes the attentive feature converter. We then utilize this feature for disparity estimation. Ablation study result in Table \ref{tab:AblationStudies}on the highlights its superior effectiveness. Consequently, this feature serves as a supervision signal for feature distillation in the \textit{Student Model}.

The \textit{Teacher Model} is trained using the ground truth disparity map, adhering to the sequential loss from \cite{RaftStereo}:
\begin{equation}
    L_{disp}^T = \sum_{i=0}^{N} \gamma^{N-i} \left\|d
    ^{gt}-\hat{d}^T_{i}\right\|,
\end{equation} 
where $i$ is the current iteration, $N$ represents the total iteration count (set to 12), and a decay weight $\gamma$ of 0.95.

\subsection{Student~:~Parallel Stereo Matching Network with Contrast-Balanced Distillation.}\label{student_model_chap}
The adaptive fusion of fog and clean features produces a powerful feature, $\widetilde{F}^{T,fusion}$, optimal for stereo matching. However, in real-world scenarios, simultaneous acquisition of both clean and hazy image pairs for inference is impractical. Therefore, we introduce a \textit{Student Model} that distills the features from the \textit{Teacher Model} as depicted in Fig.~\ref{fig:MainArchitecture}. It has a similar architecture to the \textit{Teacher Model} but can accept single domain inputs during inference.

In the training phase, the \textit{Student Model} processes paired clean and hazy images similarly, yielding initial features $F^{S,clean}$, $F^{S,fog}$ and converted ones $\widetilde{F}^{S,clean}$, $\widetilde{F}^{S,fog}$. Contrary to the \textit{Teacher Model}, which combines converted clean and fog features, the \textit{Student Model} enhances feature robustness through contrastive learning incorporated distillation. This approach maintains a balanced emphasis on fog depth hints and matching features during distillation by contrastive learning, avoiding excessive reliance on either.

In the contrastive feature distillation framework outlined in Fig.~\ref{fig:MainArchitecture}, the features $\widetilde{F}^{T,fusion}$ from the \textit{Teacher Model} is used to supervise the converted fog feature $\widetilde{F}^{S,fog}$ and clean feature $\widetilde{F}^{S,clean}$. Meanwhile, the contrastive learning $L^{cont}$ aims for feature consistency between $\widetilde{F}^{S,fog}$ and $\widetilde{F}^{S,clean}$ and to encourage difference between $\widetilde{F}^{S,fog}$ and $F^{S,fog}$. 

Intuitively, the distillation loss drives both the converted fog and clean features to closely resemble the adaptively fused features from the \textit{Teacher Model}. However, using the shared feature extractor and converter for both domains can lead to unbalanced performance, potentially making the foggy domain easier to learn and generalize badly on the clean domain. To counteract this, the triplet loss first promotes consistency between the converted clean and foggy features, ensuring balanced proficiency in both domains. Moreover, it differentiates between initial and converted fog features, preventing the converter from being trivial. preventing the converter from being trivial.

Concretely, for the feature distillation, we minimize the L1 distance between the teacher feature and the student features: 
\begin{gather}
\resizebox{.91\hsize}{!}{
   $L^{Dist} = \left\|\widetilde{F}^{T,fusion}-\widetilde{F}^{S,clean}\right\|+\left\|\widetilde{F}^{T,fusion}-\widetilde{F}^{S,fog}\right\|,$}
\end{gather}
Regarding contrastive learning, we use the Triplet Loss \cite{TripletLoss}:
\begin{gather}
\resizebox{.914\hsize}{!}{
$L^{Cont} = max(D(\widetilde{F}^{S,fog},\widetilde{F}^{S,clean})-D(\widetilde{F}^{S,fog},F^{S,fog}))+m,0),$}
\end{gather}
where the $m$ is the distance threshold which in our case equals 1.0. $D$ is the distance function defined as:
\begin{equation}
\resizebox{.70\hsize}{!}{
    $ D(F_1,F_2) = \frac{1}{C}\sum_{i=0}^{C}(\left\|F^{norm}_{1,i}-F^{norm}_{2,i}\right\|^2_2)$,}
\end{equation}
where $C$ means channel numbers and $F^{norm}$ denotes feature normalized with respect to channel dimension. The \textit{Student Model} is also supervised by the disparity loss defined as:
\begin{equation}
\resizebox{.918\hsize}{!}{
$ L_{disp}^S = \sum_{i=0}^{N} \gamma^{N-i} \left\|d
    ^{gt}-\hat{d}^{T,clean}_{i}\right\|+\sum_{i=0}^{N} \gamma^{N-i} \left\|d^{gt}-\hat{d}^{T,fog}_{i}\right\|.$
   }
\end{equation} 
The final loss of the \textit{Student Model} is the weighted summation of disparity loss and feature loss:
\begin{gather}\label{total_loss}
    L_{student} = L_{disp}^S+ \lambda_{1}L^{Dist}+\lambda_{2}L^{Cont},
\end{gather}

\section{Experiments}

\begin{table*}[!th]
\centering
\normalsize
\setlength{\tabcolsep}{1mm}
\caption{Ablation studies on the SceneFlow dataset. We report the EPE and P1-Error for both clean scenes and foggy scenes.'C' means training on clean images only, while 'F' means training on foggy images only. 'Mix' denotes using mixed clean and foggy images for training. The 'Dist' is short for distillation, and 'Cont' is short for contrastive learning. Note that the \textit{Teacher Model} requires both clean and corresponding foggy images for disparity estimation, which cannot be directly compared with other settings, marked by "*" in the table.}
\scalebox{0.96}{
\begin{tabular}{cl|cl|ccc|c|cc|cc}
\hline
\multicolumn{2}{c|}{\multirow{2}{*}{Method}} &
  \multicolumn{2}{c|}{\multirow{2}{*}{Training Data}} &
  \multicolumn{3}{c|}{Loss} &
   &
  \multicolumn{2}{c|}{EPE$\downarrow$} &
  \multicolumn{2}{c}{P1(\%)$\downarrow$} \\ \cline{5-7} \cline{9-12} 
\multicolumn{2}{c|}{} &
  \multicolumn{2}{c|}{} &
  Disp Loss &
  Triplet Loss &
  Distillation Loss &
   &
  \multicolumn{1}{c|}{Clean} &
  Foggy &
  \multicolumn{1}{c|}{Clean} &
  Foggy \\ \cline{1-7} \cline{9-12} 
\multicolumn{2}{c|}{Student-C} &
  \multicolumn{2}{c|}{Clean} &
  \checkmark &
   &
   &
   &
  \multicolumn{1}{c|}{0.628} &
  8.69 &
  \multicolumn{1}{c|}{7.3\%} &
  19.7\% \\
\multicolumn{2}{c|}{Student-F} &
  \multicolumn{2}{c|}{Foggy} &
  \checkmark &
   &
   &
   &
  \multicolumn{1}{c|}{1.710} &
  0.580 &
  \multicolumn{1}{c|}{13.7\%} &
  8.4\% \\
\multicolumn{2}{c|}{Student-Mix} &
  \multicolumn{2}{c|}{Clean + Foggy} &
  \checkmark &
   &
   &
   &
  \multicolumn{1}{c|}{0.731} &
  0.560 &
  \multicolumn{1}{c|}{8.3\%} &
  8.1\% \\ \cline{1-7} \cline{9-12} 
\multicolumn{2}{c|}{\textit{Teacher Model}} &
  \multicolumn{2}{c|}{\textit{Clean + Foggy}} &
  \checkmark &
   &
   &
   &
  \multicolumn{2}{c|}{\textit{0.352*}} &
  \multicolumn{2}{c}{\textit{5.6\%*}} \\ \cline{1-7} \cline{9-12} 
\multicolumn{2}{c|}{Student + Dist} &
  \multicolumn{2}{c|}{Clean + Foggy} &
  \checkmark &
   &
  \checkmark &
   &
  \multicolumn{1}{c|}{0.677} &
  0.499 &
  \multicolumn{1}{c|}{7.5\%} &
  7.2\% \\
\multicolumn{2}{c|}{Student + Cont} &
  \multicolumn{2}{c|}{Clean + Foggy} &
  \checkmark &
  \checkmark &
   &
   &
  \multicolumn{1}{c|}{0.613} &
  0.449 &
  \multicolumn{1}{c|}{7.0\%} &
  6.9\% \\
\multicolumn{2}{c|}{Student + Dist + Cont} &
  \multicolumn{2}{c|}{Clean + Foggy} &
  \checkmark &
  \checkmark &
  \checkmark &
   &
  \multicolumn{1}{c|}{0.552} &
  0.401 &
  \multicolumn{1}{c|}{6.5\%} &
  6.1\% \\ \hline
\end{tabular}}
\label{tab:AblationStudies}
\vspace{-3mm}
\end{table*}
\begin{figure*}[!htp]
    \centering
\scalebox{0.95}{
\includegraphics[width=1.00\linewidth]{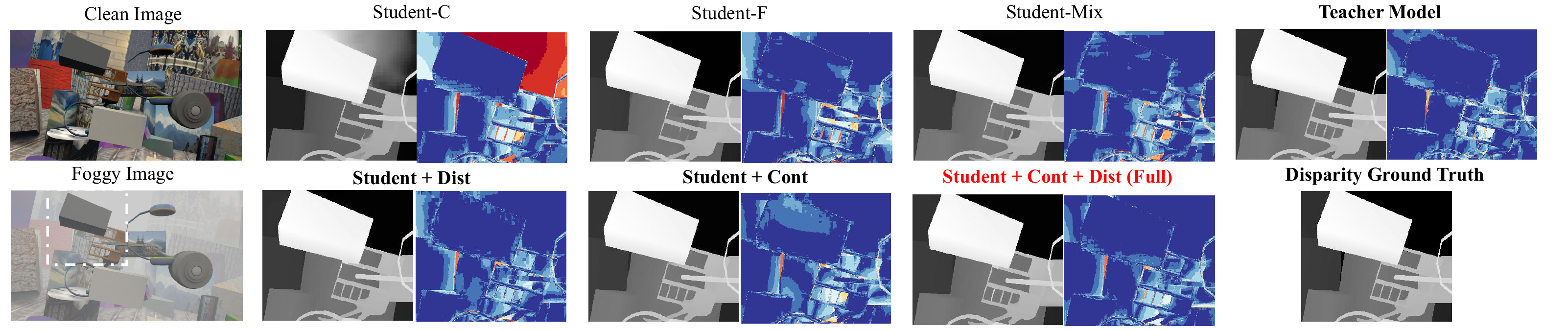}}
    \caption{Visualizations of ablation study on SceneFlow dataset. The first column is the predicted disparity for each group, while the second is the corresponding error map. We enlarged the selected part~(white bounding box) for easier viewing.}
    \label{fig:Ablation_Studies_sF}
\end{figure*}

\begin{table*}[!th]
\setlength{\tabcolsep}{1mm}
\centering
\normalsize
\caption{Quantitative comparison of \textit{CFDNet} and other methods on the SceneFlow dataset. We compared the results on clean and foggy data.~'*' represents our reimplementation results. ~\textbf{Bold: best}.}
\scalebox{0.91}{
\begin{tabular}{cl|c|c|c|ccc|c}
\hline
\multicolumn{2}{c|}{\multirow{2}{*}{Testing}} &
  \multirow{2}{*}{Metirc} &
  \begin{tabular}[c]{@{}c@{}}Standard Stereo\end{tabular} &
  Sequential &
  \multicolumn{3}{c|}{\begin{tabular}[c]{@{}c@{}}Physics-Based Joint-Learning\end{tabular}} &
  \multirow{2}{*}{\textit{CFDNet}(Ours)} \\ \cline{4-8}
\multicolumn{2}{c|}{} &
   &
RaftStereo~\cite{RaftStereo} &
FFANet*~\cite{FFANet} + RaftStereo&
  \multicolumn{1}{c|}{SDNet*~\cite{song2018deep}} &
  \multicolumn{1}{c|}{SSMDNet*~\cite{song2020simultaneous}} &
  FoggyStereo*~\cite{FoggyStereo} &
   \\ \hline
\multicolumn{2}{c|}{\multirow{2}{*}{Clean}} &
  EPE$\downarrow$ &
  0.69&
  0.76 &
  \multicolumn{1}{c|}{4.26} &
  \multicolumn{1}{c|}{3.61} &
  1.00 &
  \textbf{0.55} \\
\multicolumn{2}{c|}{} &
  3px(\%)$\downarrow$ &
  3.7\% &
  3.8\% &
  \multicolumn{1}{c|}{24.3\%} &
  \multicolumn{1}{c|}{19.8\%} &
  4.8\% &
  \textbf{3.5\%}\\ \hline
\multicolumn{2}{c|}{\multirow{2}{*}{Foggy}} &
  EPE$\downarrow$ &
  1.87 &
  1.33 &
  \multicolumn{1}{c|}{2.48} &
  \multicolumn{1}{c|}{2.08} &
  0.66 &
   \textbf{0.40} \\
\multicolumn{2}{c|}{} &
  3px(\%)$\downarrow$ &
  11.6\% &
  7.9\% &
  \multicolumn{1}{c|}{16.3\%} &
  \multicolumn{1}{c|}{13.8\%} &
  4.0\% &
   \textbf{2.9\%} \\ \hline
\end{tabular}}
\label{tab:Comparsion_On_SF_Others}
\vspace{-4mm}
\end{table*}

\begin{figure}
    \centering
\includegraphics[width=1.00\linewidth]{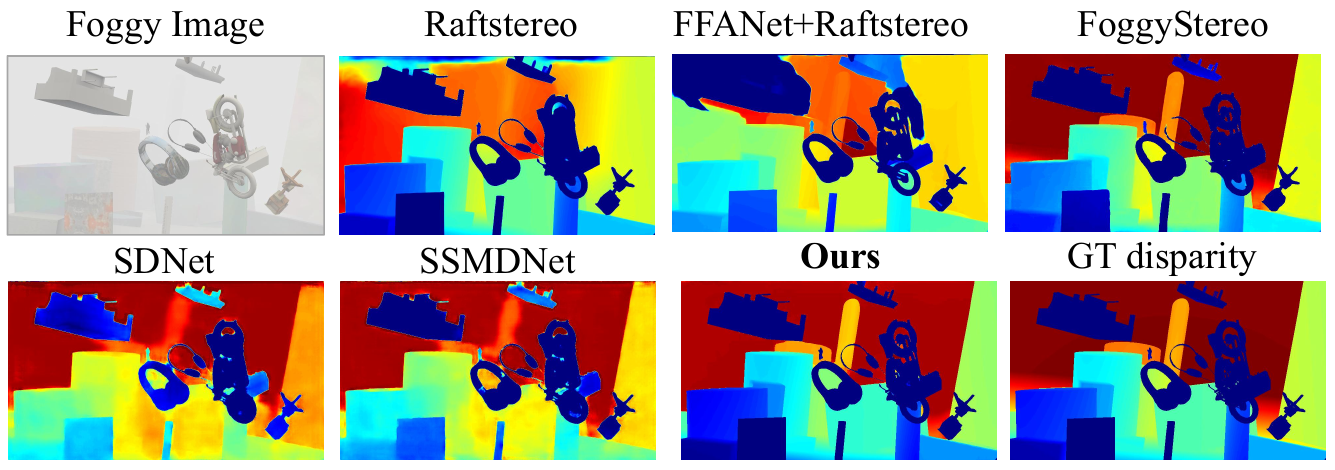}
\caption{Qualitative comparison on SceneFlow dataset foggy images with other superior works.}    \label{fig:compared_with_others}
    \vspace{-6mm}
\end{figure}

\subsection{Datasets}\label{Dataset_Introduction}
We evaluate our methods on multiple public benchmark datasets including SceneFlow~\cite{dispnetc}, KITTI~\cite{KITTIDatset}, and PixelAccurateDepth\cite{PixelAccurateDepth}. As the training of the proposed network requires pair clean and foggy images, we use the atmosphere scatter equation\cite{SingleImageVisibility} to render the foggy using the ground-truth disparity for SceneFlow and the KITTI dataset. More details about the foggy scenes synthesis can be seen in Subsection~\ref{FogImagesSynthesis}. The Sceneflow is a synthetic dataset containing 39823 stereo image pairs with random flying objects. The KITTI dataset comprises real-world scenes that have sparse ground-truth disparity captured using LiDAR. The PixelAccurateDepth\cite{PixelAccurateDepth} dataset is a real-world dataset that includes stereo images with 17 levels of artificial fogs and dense ground-truth disparity annotations.

\subsection{Foggy Images Synthesis}\label{FogImagesSynthesis}
For the foggy scenes synthesis, we use the atmosphere scattering function following the previous methods\cite{song2020simultaneous,FoggyStereo,song2018deep} using clean images and its corresponding depth map:
\begin{gather}
    I(x) = J(x)T(Z_x) + L_{\infty}(1-T(Z_x)),
\end{gather}
where $I(x)$ is the rendered foggy images, $J(x)$ is the original clean images, and $L_{\infty}$ stands for the atmospheric light of the given scenes. $T(Z_x)$ is the transmission term indicating the attenuation from an object to the camera which is commonly measured by the Beer-Lambert-Bouguer law\cite{Beer_Lambert_Law}:
\begin{align}
    T(Z_{x})= e^{-\int_{0}^{Z_{x}}\beta{(z)}dz},
\end{align}   
where the $Z_{x}$ is the given depth at location $x$, $\beta(.)$ is the attenuation coefficient which reflects the overall concentration of fog. We leverage dense ground truth disparity maps during the synthesis procedure for the Sceneflow dataset.  However, the KITTI 2012 \& 2015 datasets comprise solely sparse ground truth annotation. Therefore, we adopt a similar approach to \cite{FoggyStereo} and employ the pre-trained LEAStereo\cite{LEAStereo} model for generating pseudo-dense disparity maps, which are then employed in foggy data synthesis.
\subsection{Implementation Details}\label{ImplementationDetails}
We implemented our \textit{CFDNet} by PyTorch trained with two NVIDIA 3090 GPUs. For the SceneFlow dataset, we trained the Teacher Model for 30 epochs using a batch size
of 8 with an initial learning rate of 4e-4 following a step learning rate decay strategy. After that, we train the Student Model with the Teacher Model's intermediate feature for another 30 epochs with distilling loss and triplet loss. The $\lambda_{1}$ and $\lambda_{2}$ in Eq \ref{total_loss} are both set to 0.1 in our cases. As for the KITTI 2015\& KITTI 2012 dataset, the ground truth disparity for the benchmark test set is not available for foggy image rendering and testing. Hence, we use the KITTI 2015 training set as training data and fine-tune our SceneFlow pre-trained model with an initial learning rate of 1e-4 for 500 epochs. Then, we use the KITTI 2012 training set as test data to evaluate the performance. For the evaluation of the PixelAccurateDepth dataset, we directly used the model pre-trained on the KITTI 2015 dataset to test the real-fog generalization performances.
\subsection{Evaluation Metrics.}
For evaluation on the Sceneflow dataset, we report the standard end-point error (EPE) and the P1-value (outliers percentage) for both clean images and foggy images. As for the KITTI dataset, besides the EPE error, we follow the official evaluation and compute D1 error, which indicates the ratio of outliers with errors bigger than 3 pixels. As for the PixelAccurateDepth dataset, we follow the metrics in \cite{PixelAccurateDepth}.

\subsection{Ablation Studies}
We conducted ablation studies on the SceneFlow dataset with the following settings: (1) Student Model trained on clean scenes only (\textit{Student-C}). (2) Student Model trained on foggy scenes only (\textit{Student-F}). (3) Student Model trained on clean-and-fog mixed data( \textit{Student-Mix}). For the student model, we use a  simplified version of \cite{RaftStereo} as the baseline.

As shown in Table \ref{tab:AblationStudies}, Training on clean or foggy images independently yields favorable results within the respective image domains. However, a significant decline in performance is observed when transitioning to a different image domain, as exemplified by \textit{Student-C} (Clean $\xrightarrow{}$ Foggy: 0.68 $\xrightarrow{}$ 8.69) and \textit{Student-F} (Foggy $\xrightarrow{}$ Clean: 0.580 $\xrightarrow{}$ 1.171).  
Employing mixed training, as demonstrated by \textit{Student-Mix}, can partially mitigate this issue, but it is still inferior to the best performance in each image domain. The \textit{Teacher Model} using adaptively fused clean and foggy features substantially outperforms other models, achieving an EPE of 0.352. However, its prerequisite for dual inputs is impractical for inference in real-world applications.
With distillation from the \textit{Teacher Model}, our \textit{Student+Dist} using \textit{Student Model} for inference demonstrates superior results in both clean and foggy scenes compared to \textit{Student-Mix}. Moreover, the triplet loss for contrastive learning embedded in \textit{Student+Cont} further boosts robustness across both domains. Combining feature distillation and contrastive learning as our CFDNet framework, we observed peak performance, showing a 24.4\% and 28.3\% improvement over \textit{Student} for clean and foggy scenes, respectively. As illustrated in Figure~\ref{fig:Ablation_Studies_sF}, our proposed network, incorporating contrastive learning feature distillation, exhibits superior performance by producing more structural details and fewer errors,  while the baseline settings struggle to produce accurate estimations.

\subsection{Peformance Evaluation}
\noindent\textbf{Compared Methods}: We compare our proposed \textit{CFDNet} (\textit{Student Model} with full loss functions) with three other methods, including standard stereo matching: RaftStereo\cite{RaftStereo}, sequential two-stage dehazing and stereo matching: FFANet*\cite{FFANet}+RaftStereo, and Physics-based joint dehazing and stereo matching: SDNet\cite{song2018deep}, SSMDNet\cite{song2020simultaneous} and FoggyStereo\cite{FoggyStereo}. Since the SDNet, SSMDNet, and FoggyStereo's codes were unavailable, we reimplemented them with the original settings and trained them on the same data. We conducted experiments on the famous Sceneflow and KITTI datasets with synthetic foggy images as well as the PixelAccurateDepth dataset with real foggy scenes for generation evaluation.

\begin{table*}[t]
\centering 
\begin{minipage}{0.5\textwidth}
\centering
\small
\renewcommand\arraystretch{1.3}
\setlength\tabcolsep{1.5pt}
\caption{The comparisons on the KITTI 2012 dataset. ’*’ represents our reimplementation results.~\textbf{Bold: best}.}
\begin{tabular}{cc|cc|cc}
\hline
\multicolumn{2}{c|}{\multirow{2}{*}{Method}} &
  \multicolumn{2}{c|}{Clean} &
  \multicolumn{2}{c}{Foggy} \\ \cline{3-6} 
\multicolumn{2}{c|}{}                            & EPE$\downarrow$ & 3px(\%) $\downarrow$ & EPE$\downarrow$   & 3px(\%)$\downarrow$  \\ \hline
\multicolumn{1}{c|}{\begin{tabular}[c]{@{}c@{}}Standard Stereo\end{tabular}} &
  RaftStereo &
  0.79 &
  \textbf{2.5\%} &
  0.84 &
  3.0\% \\ \hline
\multicolumn{1}{c|}{Sequential} & FFANet*+RaftStereo & 0.79 & 2.6\%   & 0.83 & 2.9\%   \\ \hline
\multicolumn{1}{c|}{\multirow{3}{*}{\begin{tabular}[c]{@{}c@{}}Physics-Based\\ Joint-Learning\end{tabular}}} &
  SDNet* &
  1.66 &
  10.0\% &
  1.53 &
  9.2\% \\ \cline{2-6} 
\multicolumn{1}{c|}{}           & SSMDNet*          & 1.61 & 9.4\%   & 1.51 & 9.0\%   \\ \cline{2-6} 
\multicolumn{1}{c|}{}           & FoggyStereo*      & 0.78 & 2.7\%   & 0.75 & 2.6\%   \\ \hline
\multicolumn{2}{c|}{\textit{CFDNet}(Ours)}     & \textbf{0.75} & \textbf{2.5\%} & \textbf{0.70} & \textbf{2.1\%} \\ \hline
\end{tabular}

\label{tab:Compared_On_KITTI}
\end{minipage} \hspace{1.5em}
\begin{minipage}[p]{0.46\textwidth}
    \centering
    \resizebox{\linewidth}{!}{%
    \includegraphics[width=60mm]{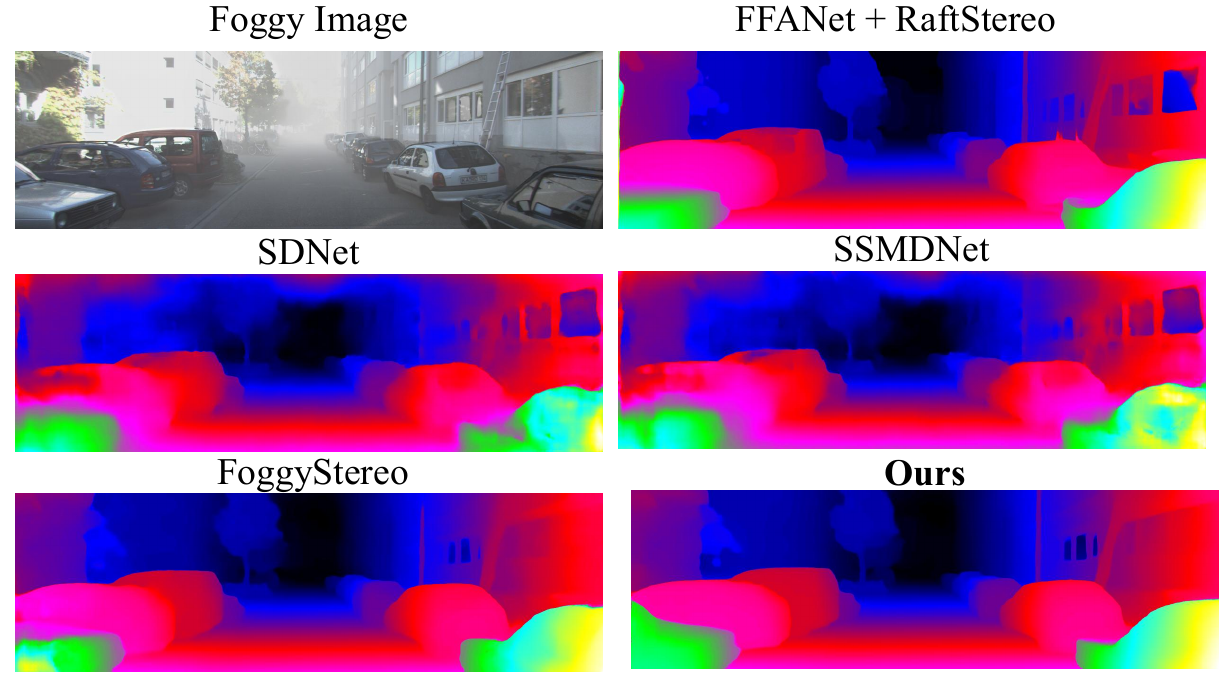} 
    }
    \caption*{Fig. 6: Visualization comparison on the KITTI 2012 dataset in the foggy scenes with other superior works.}
        \label{fig:compared_with_others_kitti}
\end{minipage} \hspace{1.5em}
\end{table*}
\begin{table*}[h]
\setlength{\tabcolsep}{1mm}
\normalsize
\centering
\caption{The generalization evaluation on real-world clean and foggy scenes of PixelAccurateDepth Dataset with KITTI pre-trained model. ’*’ represents our reimplementation results.~\textbf{Bold: Best}.}
\scalebox{0.85}{
\begin{tabular}{cc|ccllll|ccllll}
\hline
\multicolumn{2}{c|}{\multirow{2}{*}{Method}}        & \multicolumn{6}{c|}{Clean}                  & \multicolumn{6}{c}{Foggy(45m)}                    \\ \cline{3-14} 
\multicolumn{2}{c|}{}                               & RMSE$\downarrow$ & MAE$\downarrow$  & SRD$\downarrow$  & ARD$\downarrow$  & SILog$\downarrow$ & $~~\delta_{3}\uparrow$  & RMSE$\downarrow$ & MAE$\downarrow$  & SRD$\downarrow$  & ARD$\downarrow$   & SILog$\downarrow$ & $~~\delta_{3}\uparrow$  \\ \hline
\multicolumn{1}{c|}{\begin{tabular}[c]{@{}c@{}}Standard Stereo\end{tabular}} &
  RaftStereo &
  1.57 &
  0.65 &
  0.13 &
  4.02 &
  9.23 &
  99.80\% &
  2.50 &
  1.34 &
  0.36 &
  8.57 &
  13.79 &
  99.05\% \\ \hline
\multicolumn{1}{c|}{Sequential} & FFANet+RaftStereo & 1.68 & 0.68 & 0.14 & 4.10 & 9.86  & 99.71\% & 3.15 & 1.67 & 0.62 & 9.88  & 16.91 & 95.53\% \\ \hline
\multicolumn{1}{c|}{\multirow{3}{*}{\begin{tabular}[c]{@{}c@{}}Physics-Based\\ Joint-Learning\end{tabular}}} &
  SDNet* &
  1.58 &
  0.81 &
  0.14 &
  5.98 &
  10.21 &
  \textbf{99.83\%} &
  2.67 &
  1.68 &
  0.53 &
  13.01 &
  18.72 &
  98.28\% \\ \cline{2-14} 
\multicolumn{1}{c|}{}           & SSMDNet*          & 1.61 & 0.82 & 0.15 & 6.10 & 10.66 & 99.81\% & 2.60 & 1.59 & 0.49 & 12.30 & 17.84 & 98.37\% \\ \cline{2-14} 
\multicolumn{1}{c|}{}           & FoggyStereo*      & 2.27 & 0.78 & 0.18 & 6.57 & 15.40 & 99.11\% & 2.80 & 1.29 & 0.76 & 10.09 & 18.23 & 98.91\% \\ \hline
\multicolumn{2}{c|}{\textit{CFDNet}(Ours)} &
  \textbf{1.46} &
  \textbf{0.59} &
  \textbf{0.12} &
  \textbf{3.87} &
  \textbf{8.99} &
  99.82\% &
  \textbf{2.33} &
  \textbf{1.18} &
  \textbf{0.31} &
  \textbf{7.81} &
  \textbf{14.05} &
  \textbf{99.69\%} \\ \hline
\end{tabular}
}
\label{tab:Compared_On_Pixeldepth}
\vspace{-3mm}
\end{table*}
\begin{figure*}[h]
    \centering
    \scalebox{0.90}{
\includegraphics[width=1.00\linewidth]{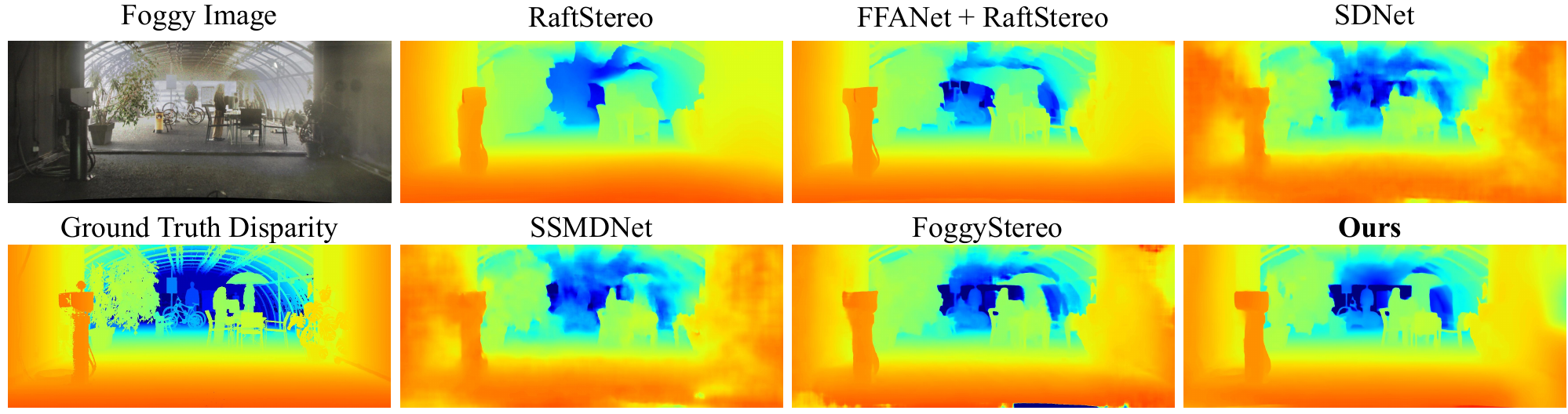}}
\caption*{Fig.~7: Generalization visualization comparison on PixelAccurateDepth dataset foggy scenes with other superior works.}  \label{fig:compared_with_pixelaccuratedepth}
    \vspace{-3mm}
\end{figure*}
\noindent\textbf{SceneFlow}: 
As shown in Table \ref{tab:Comparsion_On_SF_Others}, reimplemented methods achieved competitive or superior performance in foggy images compared to the number in original papers. Our proposed \textit{CFDNet} achieves at least 39.4\% improvement in the foggy scenes compared with the second-best FoggyStereo, and a 78.6\% improvement compared with the RAFTStereo trained on clean scenes. Furthermore, our method demonstrates superior performance on clean scenes with an EPE of 0.55, which indicates the effectiveness of our contrastive feature distillation framework for robust cross-domain performance. \\
\textbf{KITTI 2012:} As depicted in Table \ref{tab:Compared_On_KITTI}, our proposed \textit{CFDNet} outperforms existing methods when applied to foggy scenes. Additionally, our method exhibits competitive performance in clean scenes, even when compared to RaftStereo trained solely on clean images. This showcases the robustness of our contrastive feature distillation framework. Fig.~6 provides visual evidence of \textit{CFDNet}'s proficiency in generating clear and consistent disparity maps, even when operating in foggy autonomous driving scenarios.\\
\textbf{PixelAccurateDepth:} As shown in Table \ref{tab:Compared_On_Pixeldepth}, our proposed \textit{CFDNet} demonstrates outstanding generalization performance on both clean scenes and foggy scenes except for the SDNet\cite{song2018deep} with the $\delta_{3}$ metric in the clean scenes. Moreover, our method significantly outperforms existing physics-based joint dehazing and stereo-matching approaches, particularly in metrics like SRD and ARD. The visualizations in Fig. 7 offer compelling evidence of \textit{CFDNet}'s capacity for reliable disparity estimation even in foggy conditions.

\section{Conclusions}
In this paper, we have presented the \textit{CFDNet}, a novel stereo-matching network for both clean and foggy scenes. Diverging from previous methods that emphasize physical-based joint dehazing and stereo matching at the image level, our approach capitalizes on feature-level optimization using a contrastive learning incorporated feature distillation framework. This involves adaptive feature fusion for training the \textit{Teacher Model} and feature distillation to boost the \textit{Student Model}'s feature representation ability. Meanwhile, we implement a contrastive learning mechanism to balance the feature dependence on fog depth hints and clean matching features after distillation. Massive experiments on the SceneFlow, KITTI, and real-fog PixelAccurateDepth datasets have demonstrated the effectiveness of our proposed method in foggy scenes as well as our generalization ability on clean scenes.

\bibliographystyle{IEEEtran}
\bibliography{icra}

\end{document}